\documentclass[12pt]{article}
\usepackage{graphicx}
\usepackage[letterpaper,margin=1in]{geometry}
\renewenvironment{abstract}
	{\quotation}
	{\endquotation}
\date{}

\makeatletter
\renewcommand{\fnum@figure}{\textbf{Fig. \thefigure}}
\renewcommand{\fnum@table}{\textbf{Table \thetable}}
\makeatother
\usepackage{natbib}
\setcitestyle{authoryear,round}
\let\cite\citep
\usepackage{url}

\usepackage{amsmath}
\usepackage{amssymb}
\usepackage{array}
\usepackage{tabularx}
\newcolumntype{Y}{>{\raggedright\arraybackslash}X}
\def\scititle{A Bayesian framework for the uncanny valley in humanoid robot design}
\title{\bfseries \boldmath \scititle}

\author{
Shimon~Honda$^{1}$,
Rin~Shibano$^{1}$,
Hideyoshi~Yanagisawa$^{1\ast}$\and
\small$^{1}$Department of Mechanical Engineering, Graduate School of Engineering,\\
\small The University of Tokyo, Tokyo, Japan.\and
\small$^\ast$Corresponding author. Email: hide@mech.t.u-tokyo.ac.jp
}

\begin{document}
\maketitle

\begin{abstract}
The uncanny valley is a long-standing empirical rule in humanoid robot design: making robots more human-like can reduce, rather than increase, affinity. Yet existing guidelines, such as adopting robot-like appearances, avoiding excessive realism, and reducing cross-modal mismatches, remain difficult to use for algorithmic design because they are not expressed as manipulable variables. Here, we propose a hierarchical Bayesian generative model that operationalizes these guidelines as mathematical design variables. The model represents affinity toward humanoid robots as posterior-weighted negative category-conditional surprise and explains category ambiguity and perceptual mismatch as increases in surprise. It maps uncanny-valley mechanisms onto four variables: deviation from the predicted robot-category mean, inconsistency in human likeness across modalities, prediction uncertainty, and observational uncertainty. Simulations showed that category ambiguity and appearance--motion mismatch can produce affinity reductions, and that uncertainty reshapes the valley. In a human-subject experiment with robot--human morphing images, we manipulated prediction uncertainty using blurred prior robot stimuli and observational uncertainty using blurred evaluation stimuli. Increased observational uncertainty attenuated the decrease in familiarity ratings at intermediate human likeness, whereas low prediction uncertainty increased ratings for robot-like appearances. This framework turns empirical uncanny-valley heuristics into a computational basis for algorithmically evaluating and optimizing humanoid robot appearance and behavior.
\end{abstract}

\section{Introduction}

\subsection{The uncanny valley as a humanoid robot design problem}

\begin{figure}[!htbp]
\centering
\includegraphics[width=0.55\linewidth]{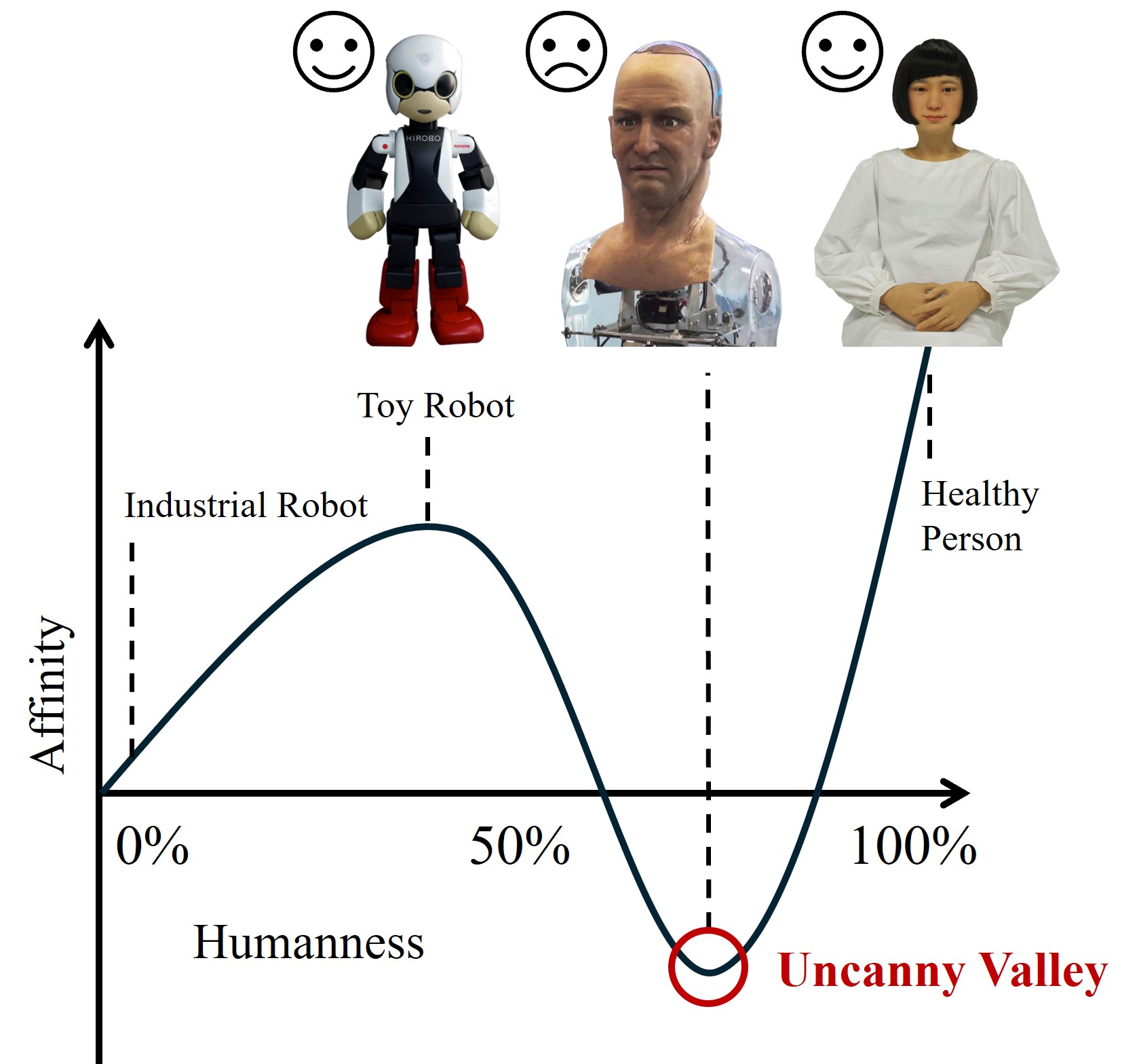}
\caption{\textbf{Schematic illustration of Mori's uncanny valley curve.}
The horizontal axis represents the human likeness of an entity, and the vertical axis represents the observer's affinity toward that entity.
As human likeness increases, affinity first increases, then sharply decreases in an intermediate region where the entity does not appear fully human, producing the uncanny valley. Affinity then increases again as the entity approaches a fully human appearance.
We reconstructed the figure based on Figure 1 in Mori\cite{Mori_2012}.
We used robot images from the Anthropomorphic roBOT (ABOT) Database\cite{Phillips_2018}.
}
\label{fig:uncanny_valley_common}
\end{figure}

For humanoid robots that interact closely with people, designers must make appearance and behavior feel familiar and acceptable. Perceived human likeness, social cues, appearance, and behavior shape trust in robots, intention to use them, emotional responses, and interaction quality\cite{van_Pinxteren_2019,Blut_2021,Song_2022,Xu_2023}. However, increasing human likeness does not always increase affinity. The uncanny valley phenomenon shows that entities that appear human-like but not fully human can sharply reduce affinity or positive impressions\cite{Mori_2012}. Figure\ref{fig:uncanny_valley_common} illustrates Mori's uncanny valley curve. Thus, humanoid robot design must consider how to avoid the uncanny valley\cite{Minh_Trieu_2023}.

Many studies have proposed design guidelines for avoiding the uncanny valley. For example, rather than aiming for a fully human appearance, researchers recommend adopting a stylized, robot-like appearance corresponding to the ``first peak'' before the uncanny valley\cite{Mori_2012,MacDorman_2025,Strait_2017,Benn_2025}. Findings that cartoon-like or non-realistic faces can elicit more trust than realistic human faces\cite{Pinney_2022}, and that excessive human likeness can reduce liking for consumer robots\cite{Kim_2019}, support this guideline. Researchers also emphasize aligning human likeness across appearance, motion, voice, touch, and other modalities to avoid cross-modal mismatches\cite{Szczepanowski_2020,Schwind_2018}. In addition, previous work highlights the need to align appearance-based expectations about competence and warmth with actual robot behavior\cite{Grazzini_2023}.

Although these guidelines have empirical support, they do not yet specify design variables mathematically. For example, the guideline of aiming for the ``first peak'' does not specify where this peak lies on the human-likeness axis. Similarly, the guideline of avoiding cross-modal mismatches does not specify how strictly designers should align modalities. As a result, designers still rely on experience and intuition when searching for appearances and behaviors that avoid uncanniness.

\subsection{From empirical design guidelines to computational mechanisms}

This problem poses an engineering challenge: infer the perceptual mechanisms behind empirical design guidelines and translate them into variables that designers can manipulate. Human-centered design emphasizes systems based on user requirements, ergonomics, and usability\cite{ISO_9241_210}. Kansei engineering offers methods for translating users' emotions and impressions into product design elements\cite{Nagamachi_1995,Schutte_2004}. More recently, HRI has increasingly used human-in-the-loop optimization to optimize robots and devices based on human responses and performance measures\cite{Slade_2024}.

Robot design widely uses mathematical models for kinematics, dynamics, control, and optimization\cite{Leone_2025,Ha_2018}. Yet few models can guide the design of appearance and behavior to avoid the uncanny valley. Researchers have also noted the lack of models that predict the uncanny valley curve in advance\cite{Wang_2015}. Moore proposed a model of the uncanny valley curve based on Bayesian category perception\cite{Moore_2012}, and Ueyama applied this model to robot therapy\cite{Ueyama_2015}. However, existing models do not primarily map uncanny-valley design guidelines onto concrete variables that designers can manipulate.

Here, we construct a hierarchical Bayesian generative model that represents affinity toward humanoid robots as posterior-weighted negative category-conditional surprise\cite{Yanagisawa_2021}. The model draws on findings that appearance--motion mismatches and deviations from human norms relate to prediction-error-like responses and negative evaluations\cite{Saygin_2011,Urgen_2018,Kawabe_2017}. The model also builds on the Bayesian brain hypothesis, which views perception as probabilistic inference about the latent causes of sensory inputs\cite{Knill_2004}, and on the free-energy principle\cite{Friston_2010}.

This formulation explains category ambiguity and perceptual mismatch, two major hypotheses for the uncanny valley, within a common framework as increases in surprise. Category ambiguity refers to a state in which an observation fits neither the human category nor the robot category sufficiently\cite{YAMADA_2012,Burleigh_2015,Cheetham_2013}. Perceptual mismatch refers to a state in which cues such as appearance, motion, and voice indicate inconsistent levels of human likeness\cite{MacDorman_2016,MacDorman_2017,Chattopadhyay_2016,Mitchell_2011,Higgins_2022}.

This formulation organizes guidelines for reducing uncanniness around four design variables. The first is the distance between observed human likeness $y$ and the predicted mean $\mu_R$ for a typical robot, $|y-\mu_R|$. This variable operationalizes the guideline of aiming for the ``first peak'' as bringing appearance closer to the predicted mean of the robot category. The second is the difference in human likeness indicated by appearance and motion, $(y_a-y_m)^2$, which represents perceptual mismatch across modalities. The third is prediction uncertainty for the robot category, $\sigma_R^2$, which represents the breadth of the observer's belief about robot appearance. The fourth is observational uncertainty, $\sigma_l^2$, which represents how precisely the observer processes appearance as sensory evidence. This mapping embeds existing design guidelines into model components and provides a basis for deriving underexplored design strategies.

Among these variables, we focus on $\sigma_R^2$ and $\sigma_l^2$, which concern observer-side uncertainty, and test their effects on familiarity in a human-subject experiment. Few uncanny valley studies have separately manipulated prediction uncertainty and observational uncertainty within the same experiment. We therefore used morphing stimuli between robot and human images, manipulated prediction uncertainty by blurring the robot image presented as a prior stimulus, and manipulated observational uncertainty by blurring the evaluation image. This design allowed us to test whether the model-predicted uncertainty effects appear in familiarity with humanoid robot appearances and in the shape of the uncanny valley.

\subsection{Contributions of the present study}

This study aims to formulate affinity perception toward humanoid robots as a Bayesian generative model composed of designer-manipulable variables and to systematize design guidelines for reducing uncanniness.

This study makes three contributions. First, we construct a Bayesian generative model that explains category ambiguity and perceptual mismatch as increases in Shannon surprise within a unified framework. This model interprets existing design guidelines, such as approaching the first peak and aligning modalities, as computational components such as $|y-\mu_R|$ and $(y_a-y_m)^2$. Second, model simulations and a human-subject experiment show that prediction uncertainty $\sigma_R^2$ and observational uncertainty $\sigma_l^2$ have distinct effects on the shape of the uncanny valley. In the experiment, increased observational uncertainty attenuated the decrease in familiarity at intermediate human likeness, and low prediction uncertainty increased familiarity for robot-like appearances. However, high prediction uncertainty did not increase familiarity at intermediate human likeness. Third, we systematize design guidelines for reducing uncanniness based on the model components. This framework organizes existing empirical guidelines and provides a theoretical basis for underexplored design strategies, such as adjusting observational precision and guiding attention. The model may also provide a mathematical basis for future algorithms that evaluate and optimize robot appearance and behavior.

\section{Results}
\subsection{Bayesian model of affinity reductions}

We model affinity as posterior-weighted negative category-conditional surprise. We first formulate a unimodal model for appearance alone and then extend it to a multimodal model that incorporates appearance and motion.

\subsubsection{Unimodal model of category ambiguity}

\begin{figure}[!htbp]
\centering
\includegraphics[width=0.95\linewidth]{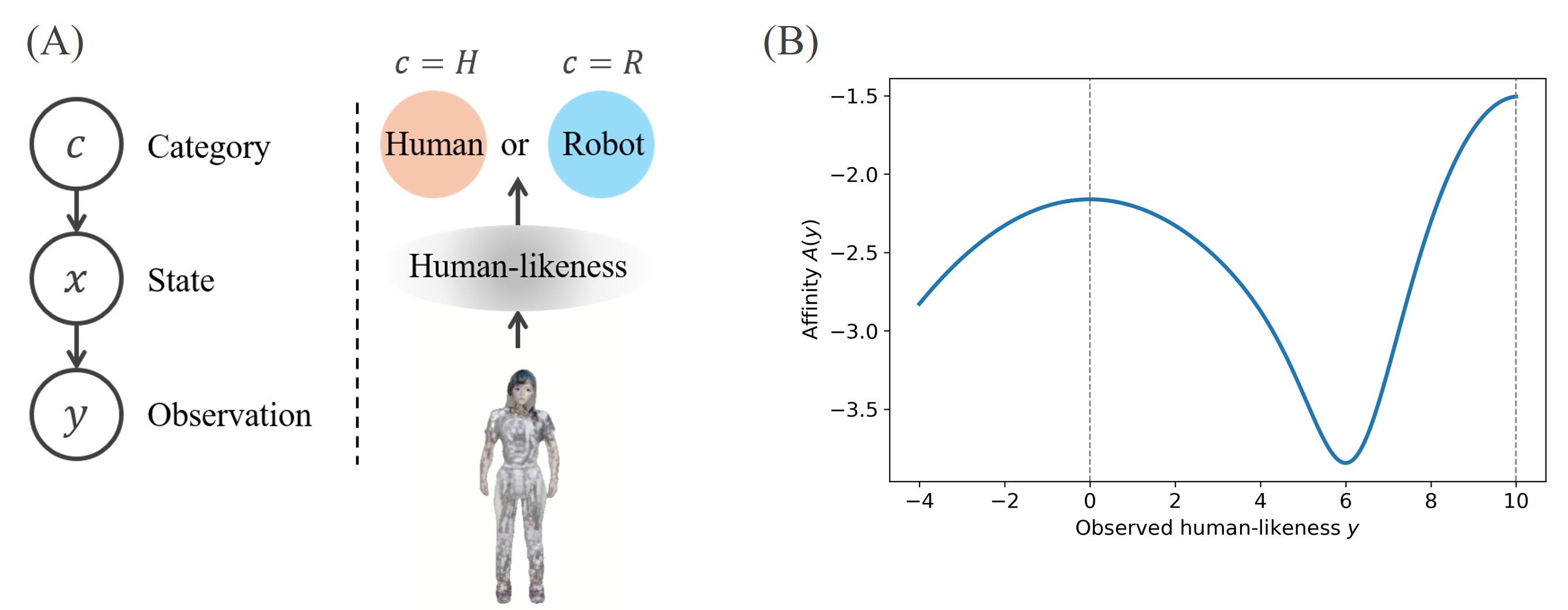}
\caption{\textbf{Explanation of category ambiguity by the unimodal model.}
(\textbf{A}) Unimodal hierarchical Bayesian generative model. The left panel shows the graphical model, in which category $c$ generates latent human likeness $x$, which then generates observation $y$. The right panel shows the corresponding interpretation in humanoid robot perception: the observer infers the entity's human likeness $x$ from appearance observation $y$ and judges whether the entity is human or robot.
(\textbf{B}) Simulated affinity measure $A(y)$ under representative parameters. The category means were $\mu_R=0$ and $\mu_H=10$, prediction uncertainties were $\sigma_R^2=10.0$ and $\sigma_H^2=1.0$, and observational uncertainty was $\sigma_l^2=2.0$. We set the prior category probabilities to $\pi_R=\pi_H=0.5$. The horizontal axis represents observed human likeness $y$, and the vertical axis represents the affinity measure $A(y)$. The vertical dotted lines indicate the category means $\mu_R$ and $\mu_H$.
}
\label{fig:unimodal_affinity}
\end{figure}

We first constructed a unimodal hierarchical Bayesian generative model for appearance. The model assumes that an observer infers the latent human likeness $x$ of an entity from an appearance observation $y$ and infers whether the entity belongs to the robot category $R$ or the human category $H$ (Fig.\ref{fig:unimodal_affinity}A). The generative model is

\begin{equation}
p(c,x,y)=p(c)p(x|c)p(y|x).
\label{eq:unimodal_genmodel}
\end{equation}

We define the observer's affinity toward an entity by how well the generative model explains the observation $y$ after category recognition. Specifically, we define the Shannon surprise for observation $y$ under category $c$ as

\begin{equation}
S_c(y)
=
-\ln p(y|c)
\label{eq:category_conditional_surprise}
\end{equation}

and define its negative as the affinity measure under category $c$:

\begin{equation}
A_c(y)
=
-S_c(y)
=
\ln p(y|c).
\label{eq:category_conditional_affinity}
\end{equation}

Thus, as category $c$ explains observation $y$ less well, conditional surprise $S_c(y)$ increases and the category-specific affinity measure $A_c(y)$ decreases. We define the final affinity measure by integrating these category-specific affinity measures according to posterior category probabilities based on $y$.

\paragraph{Baseline model.}

We assume the category-specific predictive distribution $p(x|c)=\mathcal{N}(x;\mu_c,\sigma_c^2)$ and the observation process $p(y|x)=\mathcal{N}(y;x,\sigma_l^2)$. Here, $\mu_c$ denotes the predicted mean of category $c$, $\sigma_c^2$ denotes prediction uncertainty for that category, and $\sigma_l^2$ denotes observational uncertainty. Marginalizing over the latent variable $x$ gives the predictive distribution of $y$ under category $c$:

\[
p(y|c)=\mathcal{N}(y;\mu_c,\sigma_c^2+\sigma_l^2).
\]

The affinity measure under category $c$ is therefore

\begin{equation}
A_c(y)
=
\log p(y|c)
=
-
\frac{1}{2}\log\{2\pi(\sigma_c^2+\sigma_l^2)\}
-
\frac{1}{2(\sigma_c^2+\sigma_l^2)}
\underbrace{(y-\mu_c)^2}_{\text{distance from the category mean}}.
\label{eq:category_affinity}
\end{equation}

When $c=R$, the term $(y-\mu_R)^2$ in Eq.\eqref{eq:category_affinity} represents how far the observed human likeness $y$ deviates from the observer's predicted mean $\mu_R$ for a typical robot category. This term therefore provides a design variable for interpreting the ``first peak'' as a region close to the predicted mean of the robot category.

We assume that the observer does not assign $y$ completely to either the robot category $R$ or the human category $H$. Instead, the observer integrates category-specific affinity measures according to posterior category probabilities after observation. Letting the prior category probability be $\pi_c=p(c)$, Bayes' rule gives the posterior category probability for observation $y$ as

\begin{equation}
p(c|y)
=
\frac{
p(c)p(y|c)
}{
\sum_{c' \in \{R,H\}}
p(c')p(y|c')
}
=
\frac{
\pi_c \exp\{A_c(y)\}
}{
\sum_{c' \in \{R,H\}}
\pi_{c'} \exp\{A_{c'}(y)\}
}.
\label{eq:category_posterior}
\end{equation}

Here, $A_c(y)=\log p(y|c)$. Equation\eqref{eq:category_posterior} represents how well the robot and human categories explain observation $y$.

We define the affinity measure as the category-specific affinity measures weighted by posterior category probabilities:

\begin{equation}
A(y)
=
\mathbb{E}_{p(c|y)}[A_c(y)]
=
\sum_{c \in \{R,H\}}
p(c|y)A_c(y).
\label{eq:affinity_posterior_weighted}
\end{equation}

In this formulation, when the robot category explains observation $y$ well, $p(R|y)$ becomes large and $A(y)$ mainly reflects $A_R(y)$. When the human category explains $y$ well, $p(H|y)$ becomes large and $A(y)$ mainly reflects $A_H(y)$. Near the category boundary, both posterior probabilities take intermediate values, and the model smoothly integrates the category-specific affinity measures.

Figure\ref{fig:unimodal_affinity}B shows the simulated affinity measure $A(y)$ under representative parameters. The model generated a curve in which affinity first increased with human likeness, then decreased at intermediate human likeness, and increased again near the fully human region. This result supports the category ambiguity hypothesis: affinity decreases when neither the robot nor human category sufficiently explains an intermediate appearance.

\paragraph{$\epsilon$-floor likelihood model.}

The baseline model formulates the observation process as a Gaussian likelihood. Under this assumption, surprise increases quadratically with prediction error. Thus, especially under high prediction precision, the model may overestimate surprise for observations that deviate substantially from category predictions\cite{Yanagisawa_2025}. Following our previous study\cite{Yanagisawa_2025}, we therefore also considered an $\epsilon$-floor likelihood that adds a small floor to the observation likelihood.

This model replaces only the observation process:

\begin{equation}
p_\epsilon(y|x)
=
\frac{
\mathcal{N}(y;x,\sigma_l^2)+\epsilon
}{
1+\epsilon
}.
\label{eq:epsilon_observation_likelihood}
\end{equation}

After marginalizing over $x$, the predictive distribution under category $c$ takes the form of the baseline distribution $\mathcal{N}(y;\mu_c,\sigma_c^2+\sigma_l^2)$ with the same $\epsilon$ floor. Thus, when observation $y$ deviates substantially from the category prediction and the Gaussian component approaches zero, the likelihood approaches $\epsilon/(1+\epsilon)$, which bounds surprise.

Let $A_c^\epsilon(y)$ denote the category-specific affinity measure obtained from the $\epsilon$-floor likelihood, and let $p_\epsilon(c|y)$ denote the corresponding posterior category probability. The final affinity measure becomes

\begin{equation}
A^\epsilon(y)
=
\sum_{c \in \{R,H\}}
p_\epsilon(c|y)A_c^\epsilon(y).
\label{eq:epsilon_affinity_posterior_weighted}
\end{equation}

This formulation reduces to the baseline model when $\epsilon=0$.

\subsubsection{Multimodal model of perceptual mismatch}

\begin{figure}[!htbp]
\centering
\includegraphics[width=0.95\linewidth]{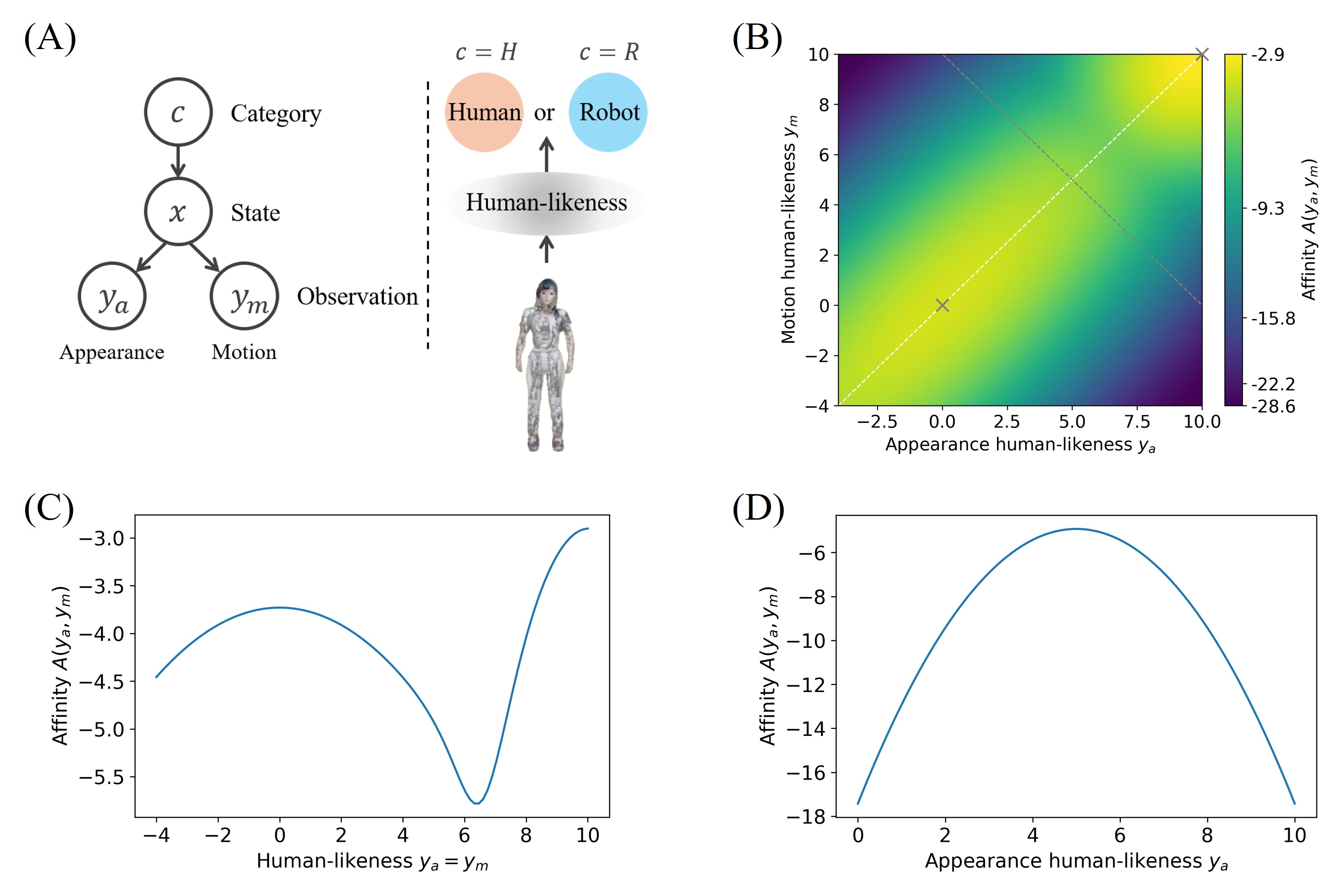}
\caption{\textbf{Explanation of category ambiguity and perceptual mismatch by the multimodal model.}
(\textbf{A}) Multimodal hierarchical Bayesian generative model. Category $c$ generates latent human likeness $x$, which then generates appearance observation $y_a$ and motion observation $y_m$.
(\textbf{B}) Heatmap of the affinity measure $A(y_a,y_m)$ for appearance human likeness $y_a$ and motion human likeness $y_m$. We set the parameters to $\mu_R=0$, $\mu_H=10$, $\sigma_R^2=10.0$, $\sigma_H^2=1.0$, $\sigma_a^2=2.0$, $\sigma_m^2=2.0$, and $\pi_R=\pi_H=0.5$. Gray crosses indicate the category means $(\mu_R,\mu_R)$ and $(\mu_H,\mu_H)$. The white dotted line indicates $y_a=y_m$, and the gray dotted line indicates $y_a+y_m=10$.
(\textbf{C}) Cross-section along $y_a=y_m$. Because appearance and motion match in human likeness, this section shows the affinity reduction mainly arising from category ambiguity, as in the unimodal model.
(\textbf{D}) Cross-section along $y_a+y_m=10$. At the center, $y_a$ and $y_m$ match, whereas the appearance--motion mismatch increases toward both ends. As a result, affinity decreases due to perceptual mismatch across modalities.
}
\label{fig:multimodal_affinity}
\end{figure}

We next extended the unimodal model to a multimodal model that incorporates appearance and motion. In the multimodal model, the observer obtains two types of sensory evidence: an appearance observation $y_a$ and a motion observation $y_m$. We assume that a common latent variable, the entity's human likeness $x$, generates both observations (Fig.\ref{fig:multimodal_affinity}A). The generative model is

\begin{equation}
p(c,x,y_a,y_m)
=
p(c)p(x|c)p(y_a|x)p(y_m|x).
\label{eq:multimodal_genmodel}
\end{equation}

We assume the category-specific predictive distribution $p(x|c)=\mathcal{N}(x;\mu_c,\sigma_c^2)$ and the observation processes $p(y_a|x)=\mathcal{N}(y_a;x,\sigma_a^2)$ and $p(y_m|x)=\mathcal{N}(y_m;x,\sigma_m^2)$. Here, $\mu_c$ denotes the predicted mean of category $c$, $\sigma_c^2$ denotes prediction uncertainty for that category, $\sigma_a^2$ denotes observational uncertainty for appearance, and $\sigma_m^2$ denotes observational uncertainty for motion. Under these assumptions, the affinity measure under category $c$ is

\begin{equation}
\begin{aligned}
A_c(y_a,y_m)
=
&-\log(2\pi)
-
\frac{1}{2}\log D_c \\
&-
\frac{1}{2D_c}
\left\{
\underbrace{(y_a-y_m)^2}_{\substack{\text{mismatch}\\\text{across modalities}}}
\sigma_c^2
+
\underbrace{(y_a-\mu_c)^2}_{\substack{\text{distance from}\\\text{the category mean (appearance)}}}
\sigma_m^2
+
\underbrace{(y_m-\mu_c)^2}_{\substack{\text{distance from}\\\text{the category mean (motion)}}}
\sigma_a^2
\right\}.
\end{aligned}
\label{eq:multimodal_affinity}
\end{equation}

where

\[
D_c
=
\sigma_a^2\sigma_m^2
+
\sigma_a^2\sigma_c^2
+
\sigma_m^2\sigma_c^2.
\]

The final affinity measure weights category-specific affinity measures by posterior category probabilities after observation:

\begin{equation}
A(y_a,y_m)
=
\sum_{c \in \{R,H\}}
p(c|y_a,y_m)A_c(y_a,y_m).
\label{eq:multimodal_affinity_posterior_weighted}
\end{equation}

Here, $p(c|y_a,y_m)$ extends $p(c|y)$ in the unimodal model to the multivariate case using the category-specific affinity measure $A_c(y_a,y_m)$.

The term $(y_a-y_m)^2$ in Eq.\eqref{eq:multimodal_affinity} represents inconsistency in human likeness between appearance and motion. Thus, when the appearance is human-like but the motion is robot-like, or vice versa, this term increases and the category-specific affinity measure $A_c(y_a,y_m)$ decreases.

Figure\ref{fig:multimodal_affinity}B shows a heatmap of the affinity measure $A(y_a,y_m)$ in the multimodal model. Affinity was high when appearance human likeness $y_a$ and motion human likeness $y_m$ were both robot-like or both human-like. In contrast, affinity decreased when appearance and motion differed substantially in human likeness.

Figure\ref{fig:multimodal_affinity}C shows a cross-section along the diagonal line $y_a=y_m$. Because appearance and motion match in human likeness along this section, the affinity reduction arises mainly from category ambiguity. Figure\ref{fig:multimodal_affinity}D shows a cross-section along the line $y_a+y_m=10$. This condition holds the average human likeness of appearance and motion constant while varying only their difference. Affinity decreased as the inconsistency between appearance and motion increased.

Together, these results show that the multimodal model predicts a valley from category ambiguity when appearance and motion are consistent, and an affinity reduction from perceptual mismatch when they are inconsistent.

\subsubsection{Uncertainty effects on affinity}

\begin{figure}[!htbp]
\centering
\includegraphics[width=0.95\linewidth]{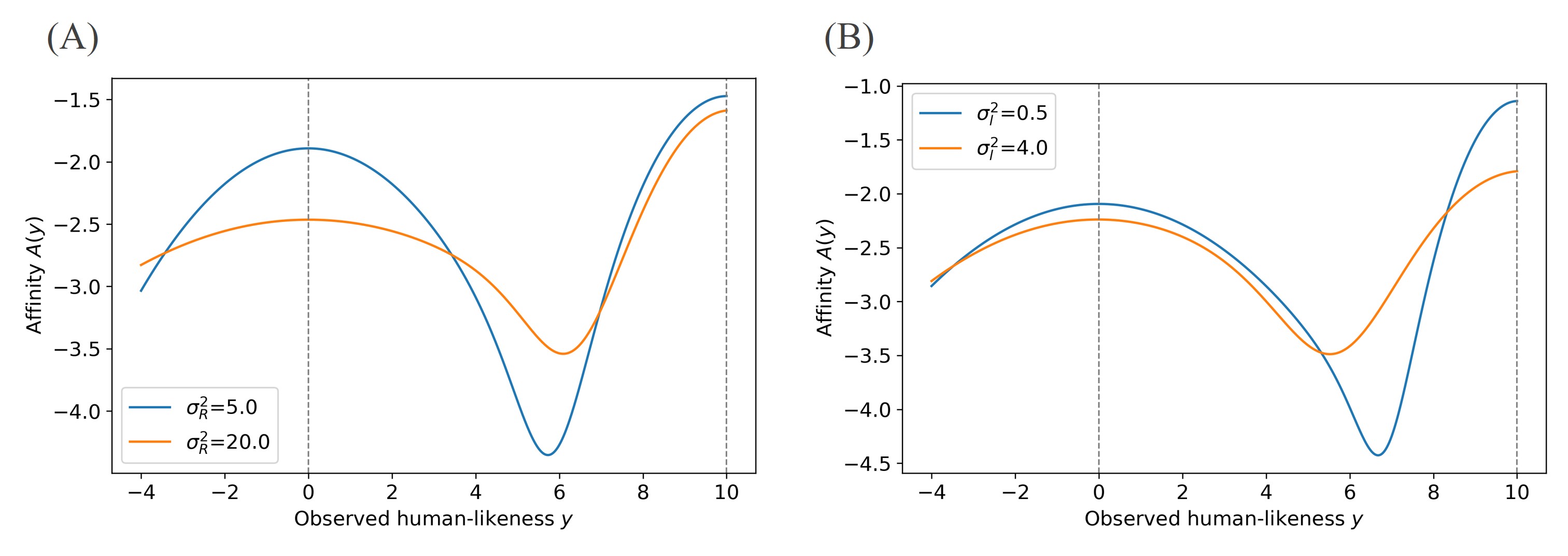}
\caption{\textbf{Effects of prediction uncertainty and observational uncertainty on the affinity curve.}
(\textbf{A}) Effects of varying prediction uncertainty for the robot category, $\sigma_R^2$. With small $\sigma_R^2$, affinity is high near the robot-category mean $\mu_R$ but sharply decreases in the intermediate region away from this mean. With large $\sigma_R^2$, the robot-side curve becomes flatter, and the decrease in affinity in the intermediate region becomes shallower.
(\textbf{B}) Effects of varying observational uncertainty, $\sigma_l^2$. With small $\sigma_l^2$, affinity substantially decreases near the category boundary, producing a deep valley. With large $\sigma_l^2$, the affinity curve becomes smoother, and the decrease in affinity at intermediate human likeness weakens.
In both panels, the horizontal axis represents observed human likeness $y$, and the vertical axis represents the affinity measure $A(y)$. The gray dotted lines indicate the robot-category mean $\mu_R$ and the human-category mean $\mu_H$.
}
\label{fig:uncertainty_effects}
\end{figure}

Among the four design variables, we next examined how prediction uncertainty for the robot category, $\sigma_R^2$, and observational uncertainty, $\sigma_l^2$, alter the affinity curve. Figure\ref{fig:uncertainty_effects} shows affinity curves obtained by varying $\sigma_R^2$ and $\sigma_l^2$ in the unimodal baseline model. These simulations yielded four theoretical predictions for the human-subject experiment.

\textbf{TH1-1: For robot-like appearances, lower prediction uncertainty increases affinity.}
Varying prediction uncertainty for the robot category, $\sigma_R^2$, mainly changed the robot-side curve (Fig.\ref{fig:uncertainty_effects}A). When $\sigma_R^2$ was small, appearances close to $\mu_R$ produced high affinity. Thus, for robot-like appearances, lower prediction uncertainty for the robot category should increase affinity.

\textbf{TH1-2: For intermediate human likeness, higher prediction uncertainty increases affinity.}
In contrast, when $\sigma_R^2$ was small, affinity sharply decreased as appearance moved away from $\mu_R$. When $\sigma_R^2$ was large, the robot-side curve became flatter, and affinity decreased more gradually for appearances slightly distant from the robot category (Fig.\ref{fig:uncertainty_effects}A). Thus, for appearances with intermediate human likeness, higher prediction uncertainty for the robot category should increase affinity.

\textbf{TH2: Higher observational uncertainty increases affinity.}
Varying observational uncertainty, $\sigma_l^2$, changed the sharpness of the overall affinity curve (Fig.\ref{fig:uncertainty_effects}B). When $\sigma_l^2$ was small, affinity substantially decreased near the category boundary, producing a deep valley. When $\sigma_l^2$ was large, this decrease weakened and the valley became shallower. Thus, higher observational uncertainty should increase affinity.

\textbf{TH3: Higher observational uncertainty attenuates the decrease in affinity with respect to human likeness.}
Under large $\sigma_l^2$, the affinity curve became smoother, and the decrease in affinity at intermediate human likeness weakened (Fig.\ref{fig:uncertainty_effects}B). Thus, higher observational uncertainty should attenuate the decrease in affinity with respect to human likeness.

These simulation results yielded theoretical hypotheses on prediction uncertainty (TH1-1 and TH1-2) and observational uncertainty (TH2 and TH3). These THs concern the model's affinity measure $A(y)$. In the next section, we map these model-based predictions onto experimentally manipulable stimulus conditions and test them as experimental hypotheses.

\subsection{Experiment on uncertainty effects}

\subsubsection{Experimental conditions and hypotheses}

\begin{figure}[!htbp]
\centering
\includegraphics[width=0.95\linewidth]{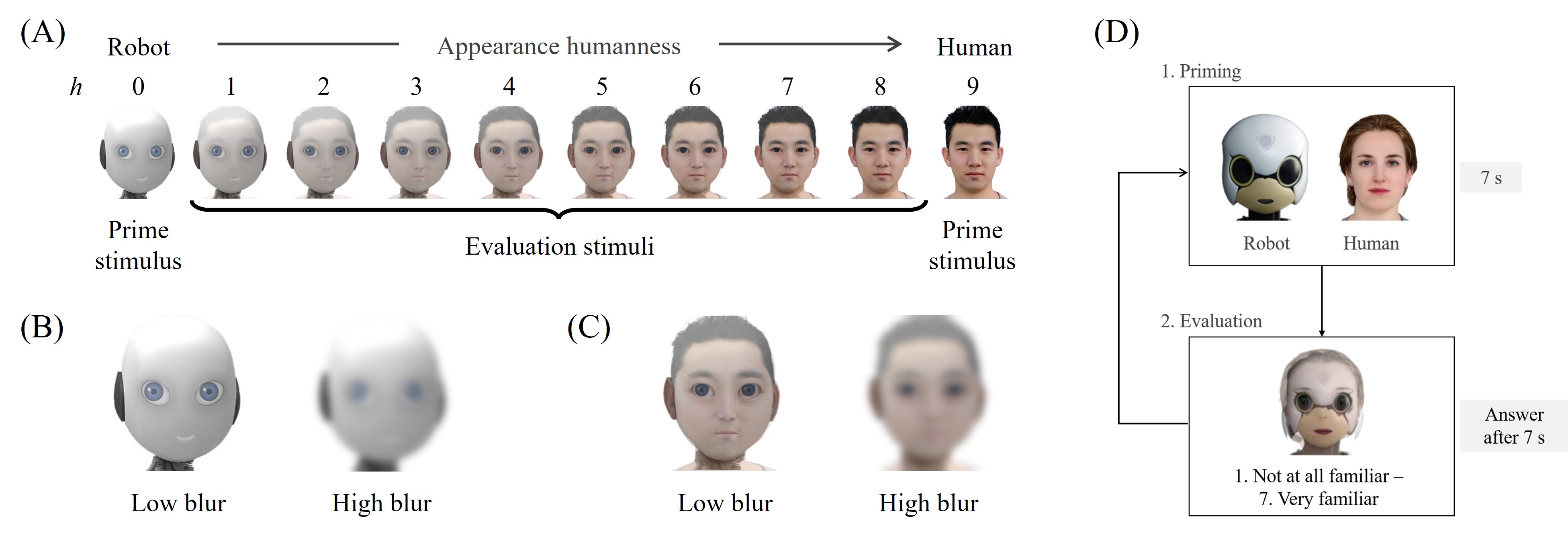}
\caption{\textbf{Example stimuli, experimental conditions, and trial procedure in the unimodal experiment.}
(\textbf{A}) Example morphing stimuli. We constructed a 10-level appearance human-likeness sequence, with the robot (Roboy) set to 0 and the human image (Japanese male) set to 9. We used the endpoint stimuli as prior stimuli and the intermediate images from 1 to 8 as evaluation stimuli.
(\textbf{B}) Blur manipulation for the prior stimulus (robot image). We used low- and high-blur conditions to manipulate prediction uncertainty.
(\textbf{C}) Blur manipulation for the evaluation stimulus. We used low- and high-blur conditions to manipulate observational uncertainty.
(\textbf{D}) Trial procedure. After viewing the prior stimulus for 7 s, participants viewed the evaluation stimulus and rated its familiarity after 7 s.
}
\label{fig:unimodal_experiment_design}
\end{figure}

We next tested the theoretical hypotheses from the unimodal baseline model in an experiment that manipulated appearance human likeness. We mapped prediction uncertainty $\sigma_R^2$ and observational uncertainty $\sigma_l^2$ onto image-blur conditions. Specifically, we used the blur level of the prior robot stimulus to manipulate prediction uncertainty $\sigma_R^2$ for the robot category, and the blur level of the evaluation stimulus to manipulate observational uncertainty $\sigma_l^2$ (Fig.\ref{fig:unimodal_experiment_design}). Because the prior robot image provided category-level information about the robot endpoint, blurring this image reduced the precision of the robot-category prior rather than the sensory precision of the subsequent evaluation stimulus. Participants rated the familiarity of each evaluation stimulus. Thirty-three adults participated. We determined the sample size based on feasibility and on the within-participant design used in previous uncanny valley studies. We excluded no participants from the analysis.

Table\ref{tab:experimental_hypotheses_uncanny} summarizes the theoretical hypotheses (THs), experimental manipulations, and experimental hypotheses (EHs). The THs concern the model's affinity measure $A(y)$, whereas the EHs test these predictions as familiarity ratings.

\begin{table}[!htbp]
\centering
\caption{\textbf{Experimental hypotheses.}
Hypotheses of the unimodal appearance experiment based on the simulation results.}
\label{tab:experimental_hypotheses_uncanny}
\scriptsize
\setlength{\tabcolsep}{2pt}
\renewcommand{\arraystretch}{0.85}
\begin{tabular}{p{0.10\linewidth}p{0.34\linewidth}p{0.22\linewidth}p{0.24\linewidth}}
\hline
Symbol & Experimental hypothesis & Experimental manipulation & Corresponding theoretical hypothesis \\
\hline
EH1-1
& For robot-like appearances, weak blur of the prior stimulus yields higher familiarity ratings than strong blur.
& Blur applied to the prior robot image. The low-blur condition corresponds to low prediction uncertainty, and the high-blur condition corresponds to high prediction uncertainty.
& For robot-like appearances, lower prediction uncertainty for the robot category increases affinity (TH1-1). \\
\hline
EH1-2
& For appearances with intermediate human likeness, strong blur of the prior stimulus yields higher familiarity ratings than weak blur.
& Blur applied to the prior robot image. The high-blur condition corresponds to high prediction uncertainty.
& For appearances with intermediate human likeness, higher prediction uncertainty for the robot category increases affinity (TH1-2). \\
\hline
EH2
& Strong blur of the evaluation stimulus yields higher familiarity ratings than weak blur.
& Blur applied to the evaluation stimulus. The low-blur condition corresponds to low observational uncertainty, and the high-blur condition corresponds to high observational uncertainty.
& Higher observational uncertainty increases affinity (TH2). \\
\hline
EH3
& Strong blur of the evaluation stimulus attenuates the decrease in familiarity ratings at intermediate human likeness compared with weak blur.
& Interaction between blur applied to the evaluation stimulus and appearance human likeness.
& Higher observational uncertainty attenuates the decrease in affinity with respect to human likeness (TH3). \\
\hline
\end{tabular}
\end{table}

\subsubsection{Effects on familiarity ratings}

\begin{figure}[!htbp]
\centering
\includegraphics[width=0.95\linewidth]{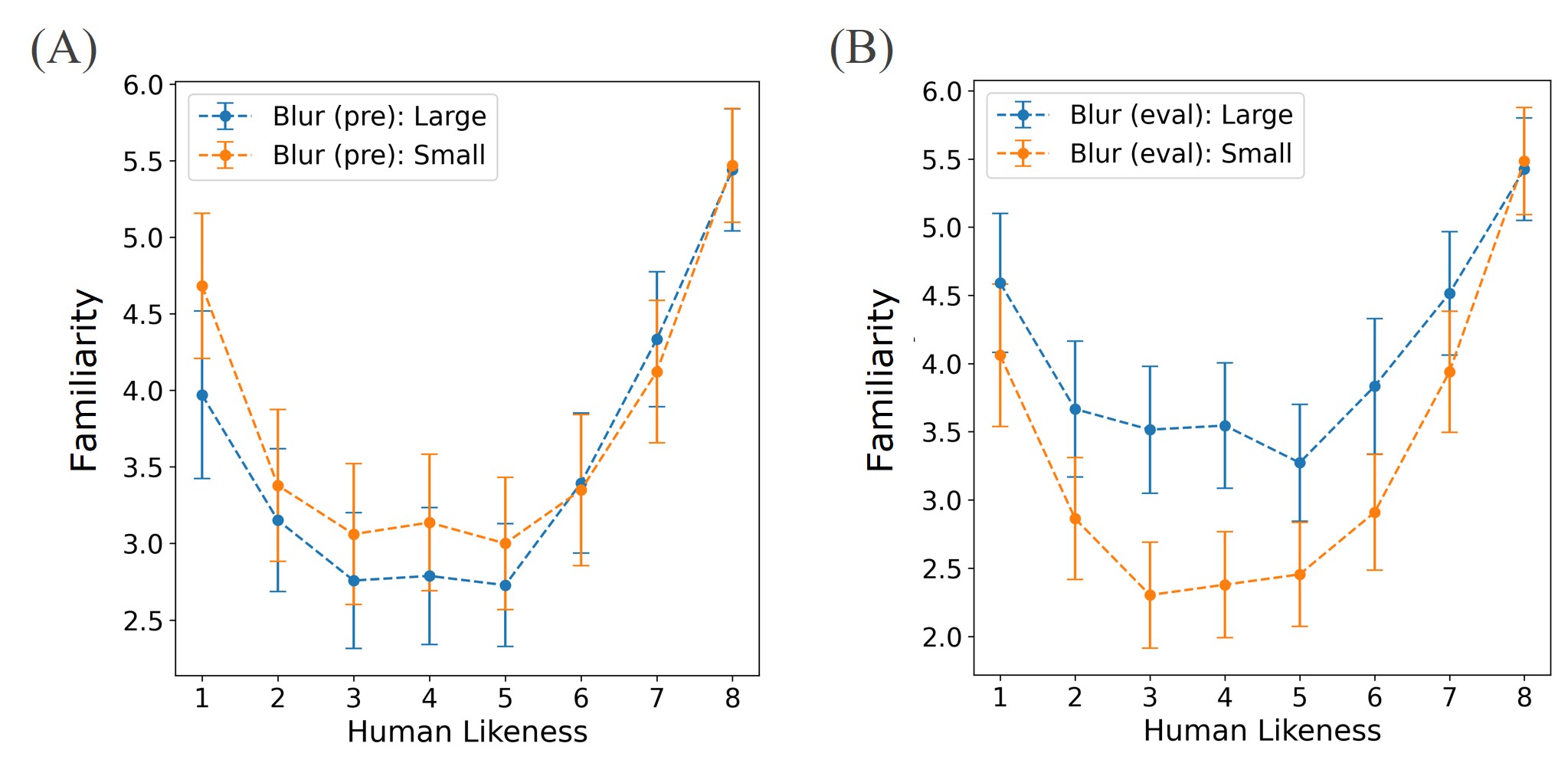}
\caption{\textbf{Familiarity ratings in the unimodal appearance experiment.}
(\textbf{A}) Difference between the low- and high-blur conditions for the prior robot stimulus. The blur manipulation for the prior stimulus corresponds to prediction uncertainty.
(\textbf{B}) Difference between the low- and high-blur conditions for the evaluation stimulus. The blur manipulation for the evaluation stimulus corresponds to observational uncertainty.
The horizontal axis represents appearance human likeness, and the vertical axis represents familiarity ratings. Error bars represent 95\% confidence intervals across participants ($n=33$).
}
\label{fig:unimodal_experiment_results}
\end{figure}

Figure\ref{fig:unimodal_experiment_results} shows the familiarity ratings. Figure\ref{fig:unimodal_experiment_results}A compares the low- and high-blur conditions for the prior robot stimulus, corresponding to the manipulation of prediction uncertainty. Figure\ref{fig:unimodal_experiment_results}B compares the low- and high-blur conditions for the evaluation stimulus, corresponding to the manipulation of observational uncertainty.

Appearance human likeness significantly affected familiarity ratings ($F=43.700$, $P < 0.001$). Familiarity varied nonlinearly with appearance human likeness and decreased for stimuli with intermediate human likeness. This result supports the basic premise of the uncanny valley: affinity decreases near the category boundary between robots and humans.

For the prediction uncertainty manipulation, participants gave higher familiarity ratings when the prior stimulus had weak blur than when it had strong blur (Fig.\ref{fig:unimodal_experiment_results}A; $F=4.486$, $P = 0.034$). However, blur applied to the prior stimulus did not significantly interact with appearance human likeness ($F=1.036$, $P = 0.404$). Thus, the results supported EH1-1, which predicted higher familiarity for robot-like appearances in the low-blur prior-stimulus condition, but did not support EH1-2, which predicted higher familiarity for appearances with intermediate human likeness in the high-blur prior-stimulus condition.

For the observational uncertainty manipulation, participants gave higher familiarity ratings when the evaluation stimulus had strong blur than when it had weak blur (Fig.\ref{fig:unimodal_experiment_results}B; $F=53.187$, $P < 0.001$). This result supports EH2, which predicted that higher observational uncertainty would increase familiarity ratings. Blur applied to the evaluation stimulus also significantly interacted with appearance human likeness ($F=2.108$, $P = 0.040$). Simple main-effect tests showed that the high-blur evaluation-stimulus condition significantly increased familiarity ratings for stimuli with human likeness levels from $h=2$ to $h=6$ ($h=2$: $F=6.031$, $P = 0.015$; $h=3$: $F=20.394$, $P < 0.001$; $h=4$: $F=20.933$, $P < 0.001$; $h=5$: $F=9.493$, $P = 0.003$; $h=6$: $F=8.989$, $P = 0.003$). We found no significant differences for $h=1$, $h=7$, or $h=8$.

Taken together, the observational uncertainty manipulation shown in Fig.\ref{fig:unimodal_experiment_results}B supported EH2 and EH3. The prediction uncertainty manipulation shown in Fig.\ref{fig:unimodal_experiment_results}A supported EH1-1 but not EH1-2. A supplementary $\epsilon$-floor model analysis showed that this model attenuated, but did not eliminate, the predicted reversal of prediction-uncertainty effects (fig. \ref{fig:model_comparison}). We return to this discrepancy in the Discussion.

\section{Discussion}

\subsection{Design variables for reducing uncanniness}

Our experiment supported three of the four hypotheses, except EH1-2. Increased observational uncertainty attenuated the decrease in familiarity at intermediate human likeness, and low prediction uncertainty increased familiarity for robot-like appearances.

Below, we organize design guidelines for reducing uncanniness according to the four design variables introduced in the Introduction. In our model, guidelines discussed in previous studies, such as approaching the first peak, aligning modalities, and avoiding expectation mismatch, correspond to operations on distinct computational components. Table\ref{tab:design_implications_uncanny_extended} summarizes the design variables, design guidelines, and model-based interpretations for reducing uncanniness.

\begin{table}[!htbp]
\centering
\caption{\textbf{Design variables for reducing uncanniness.}
The table summarizes model terms, design guidelines, examples, and supporting studies.}
\label{tab:design_implications_uncanny_extended}
\scriptsize
\setlength{\tabcolsep}{3pt}
\renewcommand{\arraystretch}{0.90}
\sloppy

\begin{tabularx}{\textwidth}{
p{0.18\textwidth}
X
p{0.12\textwidth}
p{0.34\textwidth}
}
\hline
Design variable (model term) & Design guideline & Examples & Supporting studies \\
\hline

Distance from robot-category mean
($|y-\mu_R|$)
&
Design appearances close to what observers expect as ``robot-like.''
&
Pepper, NAO\cite{Pandey_2018,Gelin_2017}
&
First-peak guidelines\cite{Mori_2012,MacDorman_2025,Strait_2017,Benn_2025}; non-realistic and stylized appearances\cite{Pinney_2022,Kim_2019}
\\
\hline

Context- or user-specific robot-category mean
($|y-\mu_R|$)
&
Make faces, expressions, and voices adjustable according to context and user.
&
Furhat\cite{Al_Moubayed_2012}
&
Adjustable robot appearance and expressions; context-dependent appearance design\cite{Schwind_2018}
\\
\hline

Mismatch across modalities
($|y_a-y_m|$)
&
Align the level of human likeness across appearance, motion, voice, and touch.
&
Geminoid HI-2\cite{Z_otowski_2015}
&
Cross-modal consistency\cite{Szczepanowski_2020,Schwind_2018}; negative responses from expectation mismatch\cite{Grazzini_2023}
\\
\hline

Prediction uncertainty
($\mu_R$, $\sigma_R^2$)
&
Promote repeated contact through routines, care, learning tasks, and personalized responses.
&
PARO, Moxie\cite{Yu_2015,Joftus_2025}
&
Reduced uncanniness through repeated exposure\cite{Z_otowski_2015,Fiolka_2024}; sustained interaction in care contexts\cite{Yu_2015}
\\
\hline

Observational uncertainty
($\sigma_l^2$)
&
Use color, surface texture, and simplified shapes to guide attention away from fine details.
&
LOVOT, PARO\cite{Yoshida_2021,Yu_2015}
&
Color use and liking\cite{Rosenthal_von_der_P_tten_2014}; texture and shape perception\cite{Todd_1997}; baby-schema features, trust, and cuteness\cite{Chen_2023,Song_2021}
\\
\hline

Local-feature weighting
(attention)
&
Distribute attention toward decorations and bodily features.
&
Hats, sunglasses, face paint, etc.\cite{Song_2025}
&
Effects of shape and decoration on gaze allocation and user experience\cite{Song_2025}
\\
\hline

\end{tabularx}
\end{table}

\subsubsection{Robot-category expectations}

The first design variable is the distance between observed human likeness $y$ and the human likeness $\mu_R$ that an observer predicts for a typical robot. The guideline of aiming for the first peak of the uncanny valley curve\cite{Mori_2012,MacDorman_2025,Strait_2017,Benn_2025} helps designers avoid full human likeness, but it does not specify which level of human likeness corresponds to that first peak. Our model interprets this guideline as bringing the appearance closer to the predicted mean $\mu_R$ of the robot category. In other words, it reframes the target ``first peak'' as the observer's representation of a typical robot appearance. Designing an appearance close to this representation should reduce uncanniness.

Observers' predictions about the robot category may also vary across users and contexts. Therefore, making faces, expressions, and voices adjustable to context and user can bring the appearance $y$ closer to that observer's prior belief. Social robots such as Furhat, which can change their face, expressions, gaze, and voice, provide an example of this design strategy\cite{Al_Moubayed_2012}.

\subsubsection{Cross-modal mismatch}

The second design variable is inconsistency in human likeness across appearance, motion, voice, touch, and other modalities. In our multimodal model, the term $(y_a-y_m)^2$ represents mismatch between appearance and motion, and the affinity measure decreases as this difference increases. Thus, even if designers make the appearance human-like, mechanical motion or voice can create perceptual mismatch and reduce affinity. Studies using Geminoid HI-2 have increased uncanniness by combining a human-like appearance with incongruent voice or jerky motion\cite{Z_otowski_2015}. Findings that uncanniness increases when visual and vocal naturalness mismatch\cite{Mitchell_2011}, as well as design suggestions emphasizing consistency across appearance, motion, sound, and touch\cite{Szczepanowski_2020,Schwind_2018}, align with our model. In addition, negative responses caused by mismatch between appearance-based expectations and actual competence or warmth\cite{Grazzini_2023} can be interpreted as inconsistency between predicted human likeness and observed behavior.

\subsubsection{Prediction uncertainty}

The third design variable is prediction uncertainty for the robot category. Prediction uncertainty represents how narrow or broad an observer's prior belief about robot appearance is. Our results suggest that, for robot-like appearances, clearer predictions about the robot category can increase familiarity.

To strengthen beliefs about a robot, designers can create mechanisms that encourage repeated contact. Previous studies show that repeated exposure reduces uncanniness toward robots\cite{Z_otowski_2015,Fiolka_2024}. For example, robots such as PARO and Moxie encourage sustained contact through care, learning tasks, and everyday conversation. Such robots can draw users' typical robot representations toward the specific robot and reduce initial prediction errors\cite{Yu_2015,Joftus_2025}.

\subsubsection{Observational uncertainty and attention}

The fourth design variable is observational uncertainty. In our experiment, increasing observational uncertainty by blurring the evaluation stimulus attenuated the decrease in familiarity at intermediate human likeness. In actual robot design, designers need not visually blur the robot itself. Instead, they can use surface texture, facial shape, color, and decoration to prevent excessive attention to fine details and thereby suppress affinity reductions.

For example, vivid colors can attract visual attention and may reduce liking for robots\cite{Rosenthal_von_der_P_tten_2014}. Thus, white or pale colors may reduce excessive attention to fine details and mitigate uncanniness. Surface texture and reflectance also affect the precision of shape perception\cite{Todd_1997}, so matte textures may prevent fine geometric details from standing out. In addition, baby-schema features increase cuteness and trust\cite{Chen_2023,Song_2021}. Rounded and soft appearances, such as those of LOVOT and PARO, can direct attention away from realistic facial details and toward tactile or animal-like familiarity\cite{Yoshida_2021,Yu_2015}. Moreover, because shape and decoration influence gaze allocation and user experience\cite{Song_2025}, hats, sunglasses, face paint, and similar features may divert attention from local features, such as the eyes and mouth, that often trigger uncanniness.

\subsection{Prediction uncertainty reversal}

EH1-1 was supported: low prediction uncertainty increased familiarity for robot-like appearances. In contrast, EH1-2 was not supported: high prediction uncertainty did not increase familiarity for appearances with intermediate human likeness. This result suggests that the baseline model may have overestimated the affinity reduction for stimuli that deviate substantially from category predictions.

\begin{figure}[!htbp]
\centering
\includegraphics[width=0.95\linewidth]{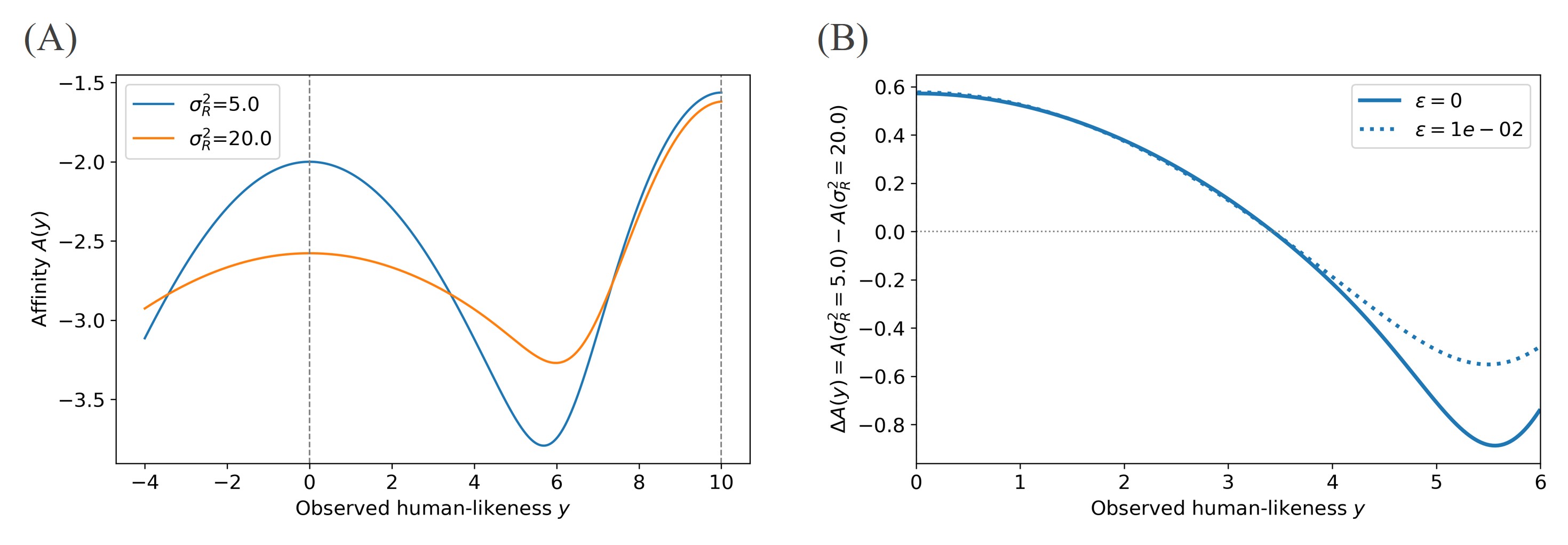}
\caption{\textbf{Comparison of prediction uncertainty effects between the baseline and $\epsilon$-floor models.}
(\textbf{A}) Affinity curves in the $\epsilon$-floor model for different levels of prediction uncertainty $\sigma_R^2$ for the robot category ($\epsilon=0.01$).
(\textbf{B}) Difference in affinity $\Delta A(y)$ between the low prediction uncertainty condition ($\sigma_R^2=5.0$) and the high prediction uncertainty condition ($\sigma_R^2=20.0$) in the baseline and $\epsilon$-floor models. The $\epsilon$-floor model attenuated the affinity difference due to prediction uncertainty in the intermediate human-likeness region.
}
\label{fig:model_comparison}
\end{figure}

In the baseline model, low prediction uncertainty for the robot category makes the likelihood of appearances far from the robot-category mean drop sharply. However, actual observers may not treat stimuli that deviate from category predictions as completely implausible. We therefore examined an $\epsilon$-floor model, which prevents likelihoods for highly deviant observations from becoming extremely small (fig. \ref{fig:model_comparison}). This model still predicted lower affinity under low prediction uncertainty in the intermediate human-likeness region, but the difference between low and high prediction uncertainty became smaller than in the baseline model. Thus, the $\epsilon$-floor model does not fully explain the lack of support for EH1-2, but it suggests that the baseline model may have overestimated the effect of prediction uncertainty and that the reversal effect may be difficult to detect experimentally.

\subsection{Limitations and future directions}

Applying our model to real robot design requires three extensions. First, future work should test the model in interactions with real robots and in real-world use contexts. This study used image stimuli and focused on appearance-based affinity evaluations. In real robot interactions, however, appearance, motion, voice, embodiment, and the surrounding environment act together\cite{Ko_2023,Tsiourti_2019}. Future studies should therefore test whether the model-derived design guidelines improve affinity and reduce uncanniness not only with image stimuli but also in real interactions and use contexts.

Second, the model should extend to personalized model fitting. Uncanniness and affinity may vary with observers' attitudes toward robots\cite{Destephe_2015,Yoganathan_2024}, interest in technology\cite{Destephe_2015}, and cultural background\cite{Castelo_2022}. The model may express these differences as parameters. For example, observers from cultures familiar with robots or observers with strong interest in technology may have lower prediction uncertainty $\sigma_R^2$ for the robot category. By estimating parameters such as $\mu_R$ and $\sigma_R^2$ for each observer, future work could connect the model to personalized robot design\cite{Gasteiger_2021,Di_Napoli_2022} and human-in-the-loop design optimization\cite{Slade_2024}.

Third, the model should incorporate context dependence. Expected appearance and behavior for humanoid robots vary with use context\cite{Dubois_Sage_2023}. In contexts where people already have clear images of humanoid robot use, such as reception, guidance, or care, observers may hold relatively clear robot-category expectations, resulting in low $\sigma_R^2$. In contrast, in unfamiliar applications, such as a humanoid robot guiding mourners at a funeral ceremony, observers may not know what appearance or behavior to expect, resulting in high $\sigma_R^2$. The predicted mean of the robot category, $\mu_R$, may also shift by context. For humanoid robots working in factories or logistics warehouses, people may expect mechanical and efficient motion as typical robot-like behavior, placing $\mu_R$ at a lower value. For reception or customer-service robots, people may expect more human-like behaviors, such as gaze shifts and nodding, placing $\mu_R$ closer to the human side. Future work should examine how designers should adjust model parameters according to use context.

\subsection{Conclusion}

This study constructed a hierarchical Bayesian generative model that represents affinity toward humanoid robots as posterior-weighted negative category-conditional surprise and explains the uncanny valley as increased surprise arising from category ambiguity and perceptual mismatch. This formulation organizes design issues such as approaching the first peak, aligning modalities, prediction uncertainty, and observational uncertainty as model variables: $|y-\mu_R|$, $(y_a-y_m)^2$, $\sigma_R^2$, and $\sigma_l^2$.

The significance of this study lies in reframing empirical design guidelines for avoiding the uncanny valley as operations on model variables that increase the affinity measure, namely negative category-conditional surprise. The model also provides a theoretical framework for deriving underexplored design strategies, such as adjusting observational precision and guiding attention. In the future, extensions to real robot interaction, personalized parameter estimation, and context-dependent evaluation functions may connect this model to design algorithms that automatically optimize humanoid robot appearance and behavior to avoid uncanniness.

\section{Materials and Methods}

\subsection{Study design}

This experiment tested whether prediction uncertainty and observational uncertainty, as defined in the model, influence familiarity ratings for humanoid robot appearances. We used a within-participant design with three factors: appearance human likeness, prediction uncertainty, and observational uncertainty. The primary outcome was the z-scored familiarity rating for each evaluation stimulus.

\subsection{Participants}

Thirty-three adults aged 18--39 years participated in the experiment (16 women and 17 men). We recruited participants who had no visual diseases or disabilities and no strong psychological resistance to humanoid robots. We recruited participants through an online participant recruitment platform\cite{JikkenBaito}.

The Research Ethics Committee of the Graduate School of Engineering, The University of Tokyo, approved this study (approval number: KE24-62). We conducted the experiment from November 18 to December 5, 2024. Before participation, we explained the study content, data to be collected, handling of personal information, voluntary participation, and the right to withdraw during the experiment without disadvantage. All participants provided written informed consent.

\subsection{Unimodal appearance stimuli}

In the unimodal experiment, we used static morphing stimuli between robot and human face images to examine affinity evaluations for humanoid robot appearance. We selected four robot images from the Anthropomorphic roBOT Database (ABOT Database)\cite{Phillips_2018}: Roboy, Lgus, JD Humanoid, and Robina. We created four human face images using Face Generator, a face-image generation service provided by Generated Photos\cite{GeneratedPhotos}: Japanese male, Japanese female, U.S. male, and U.S. female.

We morphed each robot image and human face image using Abrosoft FantaMorph 5\cite{AbrosoftFantaMorph}. We defined the appearance human-likeness level as $h$, set the robot image to $h=0$ and the human image to $h=9$, and created a 10-level morphing sequence. In the experiment, we used the eight intermediate levels, $h=1$ to $h=8$, as evaluation stimuli, excluding the robot and human endpoints.

We manipulated prediction uncertainty by changing the blur level of the prior robot image presented before the evaluation stimulus. We manipulated observational uncertainty by changing the blur level of the morphing image used as the evaluation stimulus. For both manipulations, we used low- and high-blur conditions.

We applied blur using the \texttt{cv2.GaussianBlur} function in OpenCV. In the low-blur condition, we set the kernel size to $3 \times 3$ and the standard deviation to $-1$. In the high-blur condition, we set the kernel size to $81 \times 81$ and the standard deviation to $-1$. We used the same blur settings for the prior robot image and the evaluation stimulus.

\subsection{Use of AI-assisted technologies and AI-generated stimuli}

We used Face Generator by Generated Photos to create four synthetic human face images used as source images for the morphing stimuli. The images corresponded to the following stimulus categories: Japanese male, Japanese female, U.S. male, and U.S. female. The exact prompts or interface settings used to generate these images were not retained at the time of stimulus creation. To document the stimuli used in the experiment, the generated source images, derived morphing stimuli are provided in the OSF repository. These images were used only as experimental stimuli and examples of stimuli, not as evidence or data generated by the proposed model.

AI-assisted tools were also used for limited support in manuscript editing, presentation, and code generation. Specifically, ChatGPT 5.5 was used for language refinement, wording suggestions, and formatting support during manuscript preparation, and Codex 5.5 was used to assist with generating and revising simulation, analysis, and figure-generation code. The authors reviewed and edited all AI-assisted outputs and verified the accuracy of the manuscript, citations, code, data analysis, and conclusions.

\subsection{Experimental design and procedure}

We conducted the experiment in a laboratory at the Hongo Campus of The University of Tokyo. We presented stimuli and collected responses using a web application implemented in React.js and displayed in full-screen mode on a Microsoft Surface Laptop 4. Participants sat approximately 60 cm from the screen and wore earmuffs during the experiment. We used automatic voice instructions created with VOICEVOX.

The experiment used a within-participant design that manipulated appearance human likeness, prediction uncertainty, and observational uncertainty. Appearance human likeness had eight levels ($h=1,\ldots,8$), prediction uncertainty had two levels (low/high blur of the prior robot image), and observational uncertainty had two levels (low/high blur of the evaluation stimulus). Thus, the main session comprised $8 \times 2 \times 2 = 32$ conditions.

Each participant completed three practice trials followed by 32 main trials. During the main trials, participants took a 1-min break after every 11 trials. We randomized the stimulus order for each participant. We counterbalanced the left--right positions of the human and robot images presented as prior stimuli. We also randomized the order and left--right arrangement of the subjective rating items on each trial.

In each trial, participants first viewed the human and robot images as prior stimuli for 7 s. Then, they viewed the morphing image used as the evaluation stimulus. Participants could not respond for 7 s after the evaluation image appeared, during which they observed the image. They then rated the familiarity of the presented image using a 7-point semantic differential scale. We treated the ``familiar/unfamiliar'' rating as the behavioral measure corresponding to the model's affinity measure.

\subsection{Statistical analysis}

We performed statistical analyses using R version 4.4.2. To adjust for individual differences in response tendencies, we z-scored familiarity ratings within each participant. We used the Shapiro--Wilk test to assess normality in each condition. Because some conditions violated normality, we applied an aligned rank transform (ART) to the z-scored ratings and then conducted a three-way analysis of variance.

The fixed factors were appearance human likeness (eight levels), prediction uncertainty (low/high blur of the prior stimulus), and observational uncertainty (low/high blur of the evaluation stimulus). The main tests targeted the main effects of appearance human likeness, prediction uncertainty, and observational uncertainty, as well as their interactions. When the interaction between observational uncertainty and appearance human likeness reached significance, we tested the simple main effect of observational uncertainty at each level of appearance human likeness. Error bars in the figures represent 95\% confidence intervals, and we set the significance level at $P < 0.05$.

\clearpage

%%%%%%%%%%%%%%%% REFERENCES %%%%%%%%%%%%%%%
\bibliography{references}
\bibliographystyle{apalike}

%%%%%%%%%%%%%%%% ACKNOWLEDGMENTS %%%%%%%%%%%%%%%

\section{Acknowledgments}

\paragraph*{Funding:}
S.H. was supported by The University of Tokyo project ``Advanced AI Talent Development to Lead the Next-Generation Intelligent Society (BOOST NAIS),'' funded by the Japan Science and Technology Agency (JST) through the Broadening Opportunities for Outstanding Young Researchers and Doctoral Students in Strategic Areas (BOOST) program.

\paragraph*{Author contributions:}
S.H. conceived the study. S.H., R.S., and H.Y. developed the mathematical model and interpreted the model. S.H. introduced and analyzed the $\epsilon$-floor model. S.H. and R.S. wrote the simulation code and analyzed the simulation results. S.H. prepared the figures and graphical model. S.H. and R.S. designed the human-subject experiment. R.S. created the morphing and blurred stimuli, recruited participants, conducted the experiment, and collected the data. S.H. implemented the experimental application and stimulus presentation system. S.H. and R.S. designed the statistical analysis. S.H. performed the statistical analysis. S.H. and R.S. interpreted the experimental results. S.H. wrote the original draft. S.H., R.S., and H.Y. revised the manuscript, figures, and structure. H.Y. supervised the project and provided critical intellectual input.

\paragraph*{Competing interests:}
The authors declare that they have no competing interests.

\paragraph*{Use of AI-assisted technologies:}
The authors used AI-assisted technologies to support manuscript preparation, code generation, and creation of synthetic human face stimuli. Details are provided in the Methods section. The authors reviewed all AI-assisted outputs and take full responsibility for the work.

\paragraph*{Data, code and materials availability:}
Anonymized data, analysis and simulation code, figure-generation code, and stimulus images are available at OSF: \url{https://osf.io/qc8za/?view_only=db83bf89ad534ae485f02b03b791ba80}. Stimulus images are shared in accordance with the original providers' usage conditions. No physical materials were generated in this work.

%%%%%%%%%%%%%%%% SUPPLEMENT LIST %%%%%%%%%%%%%%%

\section*{Supplementary Materials}

%%%%%%%%%%%%%%%% SUPPLEMENTARY FIGURES %%%%%%%%%%%%%%%

%%%%%%%%%%%%%%%% SUPPLEMENTARY TABLES %%%%%%%%%%%%%%%

\begin{table}[!htbp]
\centering
\small
\caption{\textbf{Three-way analysis of variance for familiarity ratings.}
The table shows the aligned-rank-transform ANOVA results for the effects of prior-stimulus blur, evaluation-stimulus blur, appearance human likeness, and their interactions.}
\label{tab:s_anova_familiarity}
\setlength{\tabcolsep}{3pt}
\begin{tabular}{p{0.34\linewidth}rrrrr}
\hline
Source & Sum of Squares & df & Mean Square & F value & P value \\
\hline
Blur (pre) & 426972.307 & 1 & 426972.307 & 4.486 & 0.034 \\
Blur (eval) & 4836813.470 & 1 & 4836813.470 & 53.187 & $<0.001$ \\
Human likeness & 22561394.636 & 7 & 3223056.377 & 43.700 & $<0.001$ \\
Blur (pre) x Blur (eval) & 117012.742 & 1 & 117012.742 & 1.225 & 0.269 \\
Blur (pre) x Human likeness & 688568.030 & 7 & 98366.861 & 1.036 & 0.404 \\
Blur (eval) x Human likeness & 1390881.303 & 7 & 198697.329 & 2.108 & 0.040 \\
Blur (pre) x Blur (eval) x Human likeness & 219810.303 & 7 & 31401.472 & 0.329 & 0.941 \\
\hline
\end{tabular}
\end{table}

\begin{table}[!htbp]
\centering
\caption{\textbf{Simple main-effect tests for evaluation-stimulus blur.}
The table shows the simple main effects of evaluation-stimulus blur at each level of appearance human likeness.}
\label{tab:s_simple_effect_eval_blur}
\setlength{\tabcolsep}{8pt}
\begin{tabular}{rrr}
\hline
Human likeness & F value & P value \\
\hline
1 & 2.538 & 0.114 \\
2 & 6.031 & 0.015 \\
3 & 20.394 & $<0.001$ \\
4 & 20.933 & $<0.001$ \\
5 & 9.493 & 0.003 \\
6 & 8.989 & 0.003 \\
7 & 3.196 & 0.076 \\
8 & 0.225 & 0.636 \\
\hline
\end{tabular}
\end{table}

\end{document}